# A novel multi-classifier information fusion based on Dempster-Shafer theory: application to vibration-based fault detection

Vahid Yaghoubi[1,2], Liangliang Cheng[1,2], Wim Van Paepegem[1], Mathias Kersemans[1]

[1]Mechanics of Materials and Structures (MMS), Ghent University, Technologiepark 46, B-9052 Zwijnaarde, Belgium.
[2]SIM M3 program, Technologiepark 48 , B-9052 Zwijnaarde, Belgium.

**Abstract**

Achieving a high prediction rate is a crucial task in fault detection. Although various classification procedures are available, none of them can give high accuracy in all applications. Therefore, in this paper, a novel multi-classifier fusion approach is developed to boost the performance of the individual classifiers. This is acquired by using Dempster-Shafer theory (DST). However, in cases with conflicting evidences, the DST may give counter-intuitive results. In this regard, a preprocessing technique based on a new metric is devised in order to measure and mitigate the conflict between the evidences. To evaluate and validate the effectiveness of the proposed approach, the method is applied to 15 benchmarks datasets from UCI and KEEL. Further, it is applied for classifying polycrystalline Nickel alloy first-stage turbine blades based on their broadband vibrational response. Through statistical analysis with different noise levels, and by comparing with four state-of-the-art fusion techniques, it is shown that that the proposed method improves the classification accuracy and outperforms the individual classifiers.

# 1 Introduction

Nowadays, employing machine learning methods for fault detection [1–3] is becoming a common task. However, the main challenge is to develop classification models with higher accuracies. In this regard, one should either develop new methods to generate proper decision boundaries e.g. discriminant analyses, neural network, support vector machines[4], and support vector data description [5], or develop proper feature selection methods, e.g., optimization-based [6] and information-based [7] methods, or combination of them, e.g. Mahalonobis classification system (MCS) [8]. Transfer learning is an alternative that could be used to enhance the performance of classification models. This could be achieved by adapting models or features from another domain to the problem at hand [9]. These approaches can provide high accuracy for a range of classification applications. But not necessarily for all applications and datasets due to the presence of different levels of noise, outliers, nonlinearities, and data redundancy [10]. To deal with this problem, researchers suggested employing an ensemble of classification models to compensate for the weaknesses and boost the strength of each classifier [11,12]. This, however, leads to two important questions: i) how to select classifiers from the pool of classifiers to keep the information and impose the diversity, and ii) how to combine their outputs to make a final decision.

To impose diversity in the ensembles, one can use different classification methods, different numbers and types of features, different training samples, etc. [13]. To combine the classifiers, several methods have been proposed in the literature. From basic elementary operations like sum, average, maximum, and minimum of the outputs [14] to more advanced forms like majority voting [15], multilayered perceptrons [16], Bayes combination [17], fuzzy integrals [18], and Dempster-Shafer theory of evidence (DST) [19–21]. The last one is of interest in this paper due to its proven advantages over other combination methods [10,13,22]. For instance, Rothe et. al. conducted an extensive investigation to analyze the performance of all the fusion methods except for the Dempster-Shafer method [10]. It was concluded that in most cases, although the fusion methods cannot outperform the best individual classifier, they decrease the sensitivity to the outliers. In contrast, in [13,22] it is shown that the Dempster-Shafer method could outperform the best individual classifier provided that the individual classifiers are independent. On the other hand, the problem of the DST method occurs when conflicting evidences come from different classifiers. In such cases, the fused output could lead to counter-intuitive results.

Several researchers targeted to solve this problem by mainly two approaches: (i) by applying different preprocessing on the evidences to reduce their possible conflict, and (ii) modifying the combination rule. Both approaches attract attention among researchers [23] but the former is at the focus of the current paper. In this regard, Deng et. al. [24] proposed to improve the basic belief assignment (BBA) based on the information extracted from the confusion matrix. In [25] the conflict between the evidences was reduced by employing Shanon's information entropy together with Fuzzy preference relations (FPR). Xiao [26] used the distance between the evidences to evaluate the support degree of the evidences (SD). Then belief entropy and FPR were employed to adjust the SD. The adjusted SD was then used as the weight for evidences prior to applying Dempster's rule of combination. In [27], a similarity between the BPAs was evaluated, and then by using it together with the belief entropy, the evidences were weighted before applying Dempster's rule of combination. In [28], a novel method has been developed to evaluate the BBAs based on the *k*-nearest neighbor algorithm. Wang et al. [29] proposed to employ both subjective and objective weight for the evidences before combination. Objective weight assesses the credibility of the evidences whereas, subjective weight evaluates the support degree of the evidences with respect to the focal elements with the largest BBAs.

The major contribution of the paper is the development of a new classifier fusion method based on Dempster-Shafer theory. The other novelties of the paper are: (i) introducing a new weighting factor based on the confusion matrix to transform the classifiers' output to the Body of evidences (BOEs), (ii) introducing a new metric for measuring the support degree of the classifiers and incorporate that in the preprocessing of the evidences. The method is then employed to enhance the performance of a vibration-based fault detection approach in detecting faulty samples when they have complex geometries.

The paper is structured as follows. In Section 2, details about the Dempster-Shafer theory and the available metrics to measure the conflicts between the evidences are presented. Section 3 elaborates the different steps in the proposed methodology, see Figure 1. In Section 4 the proposed framework is applied to multiple datasets. To benchmark the proposed framework, fifteen datasets UCI [30] and KEEL [31] repositories are selected. Further, the proposed framenwork has been applied to a vibrational dataset collected from first-stage turbine blades with complex geometry and different damage types and severities. In Section 5, the concluding remarks are presented.

## 2  Dempster-Shafer theory of evidence

Dempster-Shafer theory (DST) of evidence [19], is an important method to deal with uncertain information. It is utilized here to fuse the uncertain knowledge collected from different sources to have more concrete statistical inferences. The basic concepts of DST are provided here.

### 2.1  Basic concepts of DST

**Definition1**. Frame of discernment

Let $E_1, E_2, ..., E_K$ be a finite number of mutually exclusive and collectively exhaustive events, then the set $\mathfrak{E}$ as defined in (1) is called the Frame of discernment.

$$\mathfrak{E} = \{E_1, E_2, ..., E_K\} \qquad (1)$$

Its power set denoted by $2^{\mathfrak{E}}$ is a set of all its subsets including the null set $\varphi$ and itself $\mathfrak{E}$. $\mathcal{A}$ is called proposition if $\mathcal{A} \in 2^{\mathfrak{E}}$.

**Definition2**. Basic Belief Assignment (BBA)

For the frame of discernment $\mathfrak{E}$ BBA assigns a value in the bounded range [0, 1] to every proposition $\mathcal{A}$ of $\mathfrak{E}$ with the following conditions,

$$\begin{cases} m(\emptyset) = 0 \\ \sum_{\mathcal{A} \in 2^{\mathfrak{E}}} m(\mathcal{A}) = 1 \end{cases} \qquad (2)$$

In contrast to probability theory in which a value is assigned to individual hypotheses $E_i$, in belief theory, one can assign a value to a composite hypothesis $\mathcal{A}$, i.e. $\mathcal{A} = \{E_1, E_2\}$, without any overcommitment to either. This means some "ignorance" is associated with $\mathcal{A}$ that could lead to $m(\mathcal{A}) + m(\bar{\mathcal{A}}) \leq 1$ with $\bar{\mathcal{A}}$ as the complement of $\mathcal{A}$. This is also called mass function.

**Definition3**. Belief function and Plausibility function

For every proposition $\mathcal{A}$, belief function $0 \leq Bel \leq 1$ is defined as

$$Bel(\mathcal{A}) = \sum_{\mathcal{B} \subseteq \mathcal{A}} m(\mathcal{B}) \qquad (3)$$

And the plausibility function $0 \leq Pl \leq 1$ is defined as

$$Pl(\mathcal{A}) = \sum_{\mathcal{B} \cap \mathcal{A} \neq \emptyset} m(\mathcal{B}) = 1 - Bel(\bar{\mathcal{A}}) \qquad (4)$$

For the proposition $\mathcal{A}$, $Bel(\mathcal{A})$ is the lower limit and $Pl(\mathcal{A})$ is the upper limit of the uncertainty in the $\mathcal{A}$.

**Definition 4**. Dempster's rule of combination

For two independent bodies of evidence (BOEs) characterized by the BBAs $m_1$ and $m_2$ in the frame of discernment $\mathfrak{E}$, the Dempster's rule of combination is defined as

$$m(\mathcal{A}) = (m_1 \oplus m_2)(\mathcal{A}) = \begin{cases} \frac{m_{12}(\mathcal{A})}{1 - K} & \mathcal{A} \subset \mathfrak{E}, \mathcal{A} \neq \emptyset \\ 0 & \mathcal{A} = \emptyset \end{cases} \qquad (5)$$

in which $\oplus$ is the orthogonal sum,

$$m_{12}(\mathcal{A}) = \sum_{\mathcal{C}_1 \cap \mathcal{C}_2 = \mathcal{A}} m_1(\mathcal{C}_1) m_2(\mathcal{C}_2) \qquad (6)$$

indicates the conjunctive consensus on $\mathcal{A}$ among the propositions $\mathcal{C}_1$ and $\mathcal{C}_2$. The denominator is the normalization factor in which $K$ is a measure of conflict between the two BOEs. It is defined as,

$$K = \sum_{\mathcal{C}_1 \cap \mathcal{C}_2 = \emptyset} m_1(\mathcal{C}_1) m_2(\mathcal{C}_2), \qquad (7)$$

The combination rule can be applied to two BBAs if $K < 1$, otherwise the combination may give counterintuitive results.

The combination rule can be extended when there are several BBAs as,

$$m(\mathcal{A}) = \left(\left((m_1 \oplus m_2) \oplus m_3\right) \ldots \oplus m_n\right)(\mathcal{A}) \qquad (8)$$

The combination rules (5) and (8) provide the required foundation for classifier fusion (see in Section 3.3).

## 2.2 Conflict measures

The DST could lead to proper fusion performance provided that two conditions are satisfied, 1) $K \neq 1$, and 2) the BOEs are independent of each other. Although the former can be measured by using Dempster's rule, the latter cannot. On the other hand, the $K$ value is not enough to measure the level of conflict between the BOEs. Therefore, several measures have been defined to obtain conflict between two BOEs. To introduce some of them in the following, let $\mathcal{A}_i$ be a proposition with cardinality $|\mathcal{A}_i|$ of a BOE characterized by BBA $m$.

### 2.2.1 Belief entropy

Entropy is a mathematical tool to estimate the uncertainty present in the data. When the information is in the form of BBA, it is called belief entropy. One of the most common belief entropy is Deng entropy which is proposed by Deng [32]. It can be considered as an extension of the Shannon entropy to BBA. It is defined as

$$E_d = -\Sigma_i m(A_i) \log\left(\frac{m(A_i)}{2^{|A_i|} - 1}\right) \qquad (9)$$

The proposition with larger cardinality has larger Deng entropy which in turn, indicates more information is available in that evidence. Further, large Deng entropy for one evidence could be interpreted as more support from the other evidences.

*2.2.2 Belief divergence*

Belief divergence shows how much two BBAs are different from each other. Belief Jensen–Shannon (BJS) is a recent divergence measure proposed by Xiao [33]. For two BBAs $m_1$ and $m_2$ with $N$ mutually exclusive and exhaustive propositions in the same frame of discernment $\mathfrak{E}$, it is defined as,

$$BJS(m_1, m_2) = H\left(\frac{m_1 + m_2}{2}\right) - \frac{1}{2}H(m_1) - \frac{1}{2}H(m_2) =$$
$$= \frac{1}{2}\left[\Sigma_i m_1(A_i) \log\left(\frac{2m_1(A_i)}{m_1(A_i) + m_2(A_i)}\right) + \Sigma_i m_2(A_i) \log\left(\frac{2m_2(A_i)}{m_1(A_i) + m_2(A_i)}\right)\right] \quad (10)$$

in which $H(m_j) = -\Sigma_i m_j(A_i) \log(m(A_i))$ is the Shannon entropy.

### 2.2.3 Disagreement degree

Disagreement degree represents the extent to which one BOE is outlier comparing with the other BOEs [34]. Let $\bar{m}$ and $\bar{m}_{\sim q}$ respectively be the center of all BOEs and the center of all BOEs except for the $q$.

$$\bar{m}(A) = \frac{1}{L}\sum_{j=1}^{L} m_j(A) \quad (11)$$

$$\bar{m}_{\sim q}(A) = \frac{1}{L}\sum_{j=1, j\neq q}^{L} m_j(A) \quad (12)$$

Then, the average distance of each BOEs to these two centers are defined as,

$$SW = \frac{1}{L}\sum_{j=1}^{L} d_J(m_j, \bar{m}), \quad (13)$$

$$SW_{\sim q} = \frac{1}{L-1}\sum_{j=1, j\neq 1}^{L} d_J(m_j, \bar{m}_{\sim q}) \quad (14)$$

in which

$$d_J(m_1, m_2) = \sqrt{(m_1 - m_2)^T \mathbf{Jac}(m_1 - m_2)} \quad (15)$$

is the Jousselme's distance with $\mathbf{Jac}(A, B) = \frac{|A \cap B|}{|A \cup B|}$ as the Jaccard's weighting matrix. The disagreement degree is thus defined as,

$$m^*_{\Delta E, \sim q} = 0.5 + \frac{1}{\pi} \arctan\left(\frac{SW - SW_{\sim q}}{\sigma}\right) \qquad (16)$$

in which $\sigma$ is a parameter to adjust the effect of the $SW - SW_{\sim q}$ on the degree of disagreement. Here it is set to 2.

## 3 Proposed multi-classifier information fusion methodology

Figure 1 shows the building blocks of the proposed classifier fusion procedure. They are:

I) Classifier generation: a pool of classifiers is the outcome of this step. This is explained in Section 3.1.

II) Classifier selection: in this step, proper classifiers should be selected from the pool for combination. However, since this is not the focus of this paper, all possible combinations between the trained classifiers have been investigated, i.e. $\binom{N_c}{i}$ for $i = 1: N_c$.

III) BOE generation: in this step, the selected classifiers are transformed into the BOEs. This is explained in Section 3.2.

IV) Classifier fusion: the classifiers are combined to create a fusion model with higher accuracy than that of each individual classifier. This is extensively discussed in Section 3.3.

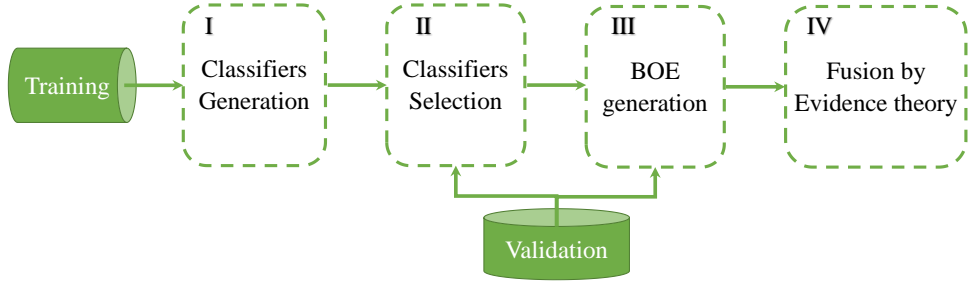

Figure 1. Flowchart of the classifier fusion method

### 3.1 Classifier generation

Here, several classifiers should be generated to create a pool of classifiers. In this regard, five classification methods are utilized. They are Linear Discriminant analysis (LDA), $k$-nearest neighbors (kNN), Support vector machines (SVM), Support vector data description (SVDD), and neural network (NN) [4]. All the classifiers are implemented using the deep learning toolbox in Matlab 2019a. The parameters associated with each method have been estimated by using 10-fold cross-validation unless otherwise mentioned explicitly.

To evaluate the performance of the trained classifiers the so-called confusion matrix, see Figure 2, is implemented. Based on this matrix, several decisive measures have been developed that are presented in Table 1. The main measure here is accuracy ($Acc$) that indicates the portion of the samples that are correctly classified, Eq. (17). The other measures are precision ($Pre$) and recall ($Rec$) which are defined for the class $k$ as Eq. (18), and Eq. (19) respectively.

$$Acc = \frac{\sum_{k=1}^{n_c} N_{kk}}{\sum_{i=1}^{n_c}\sum_{j=1}^{n_c} N_{ij}} \qquad (17)$$

$$Pre_k = \frac{N_{kk}}{\sum_{i=1}^{n_c} N_{ik}} \qquad (18)$$

$$Rec_k = \frac{N_{kk}}{\sum_{j=1}^{n_c} N_{kj}} \qquad (19)$$

|  |  | True condition | | | | |
|---|---|---|---|---|---|---|
|  |  | $c_1$ | ... | $c_k$ | ... | $c_{n_c}$ |
| Predicted condition | $\hat{c}_1$ | $N_{11}$ | ... | $N_{1k}$ | ... | $N_{1n_c}$ |
|  | ⋮ | ⋮ | | ⋮ | | ⋮ |
|  | $\hat{c}_k$ | $N_{k1}$ | ... | $N_{kk}$ | ... | $N_{kn_c}$ |
|  | ⋮ | ⋮ | | ⋮ | | ⋮ |
|  | $\hat{c}_{n_c}$ | $N_{n_c 1}$ | ... | $N_{k n_c}$ | ... | $N_{n_c n_c}$ |

Figure 2. Confusion matrix

### 3.2 Generating BOEs

Having a pool of classifiers, the next step is to transform their outputs to the BOEs. For this purpose, let $\hat{Y}_i = [\hat{y}_{i,1}, \dots, \hat{y}_{i,k}, \dots, \hat{y}_{i,n_c}] \in \mathbb{R}^{(n_s \times n_c)}$ with $n_s$ and $n_c$ as the number of samples and classes, be the output of the classifier $C_i$, $i = 1,2, \dots, N_c$. It should be emphasized that the output of the classifications are real numbers indicating the extent to which a sample belongs to the classes. The BOEs with BBA $m_i$ in the frame of discernment $\mathfrak{E} = \{E_1, \dots, E_k, \dots, E_{n_c}\}$ can then be obtained as

$$m_i(E_k) = \boldsymbol{w}_i \otimes \hat{\boldsymbol{y}}_{i,k}, \qquad (20)$$

$$m_i(\mathfrak{E}) = 1 - \sum_{k=1}^{n_c} \boldsymbol{w}_i \otimes \hat{\boldsymbol{y}}_{i,k} \qquad (21)$$

in which $m_i(\mathfrak{E})$ is the ignorance associated with the classifier $C_i$, $\otimes$ is the Kronecker product, and $\boldsymbol{w}_i \in \mathbb{R}^{1 \times n_c}$ is a weighting factor that can be obtained based on the confusion matrix (see Figure 2). Different weightings are presented in Table 1. The $\boldsymbol{w}_0 = \boldsymbol{1} \in \mathbb{R}^{1 \times n_c}$, as a vector of 1's, is the unweighted version of the BOE generation technique. The weighting $w_1 - w_4$ have been collected from the literature [24,35,36]. The last weighting, i.e. $w_5$, is introduced here as the combination of overall accuracy and class precisions by applying the Dempster rule of combination as follows,

$$w_5 = (Acc \otimes \mathbf{1}) \oplus [Pre_1, \ldots, Pre_k, \ldots, Pre_{n_c}] \tag{22}$$

The efficacy of these weightings in improving the performance of the proposed fusion technique will be investigated in the following sections.

Table 1. Weighting factors

| Response | Weight | Formulation | Description | | Reference |
|---|---|---|---|---|---|
| $P_0$ | $w_0$ | 1 | Unweighted form | | --- |
| $P_1$ | $w_1$ | $Acc \otimes \mathbf{1}$ | $Acc =$ | $\dfrac{\sum_{k=1}^{n_c} N_{kk}}{\sum_{i=1}^{n_c} \sum_{j=1}^{n_c} N_{ij}}$ | |
| $P_2$ | $w_2$ | $[Pre_1, \ldots, Pre_k, \ldots, Pre_{n_c}]$ | $Pre_k =$ | $\dfrac{N_{kk}}{\sum_{i=1}^{n_c} N_{ik}}$ | [35] |
| $P_3$ | $w_3$ | $[Rec_1, \ldots, Rec_k, \ldots, Rec_{n_c}]$ | $Rec_k =$ | $\dfrac{N_{kk}}{\sum_{j=1}^{n_c} N_{kj}}$ | [36] |
| $P_4$ | $w_4$ | $w_2 \oplus w_3$ | ---- | | [24] |
| $P_5$ | $w_5$ | $w_1 \oplus w_2$ | ---- | | |

### 3.3 New DST-based Classifier fusion

In this section, a new method for combining classifiers is proposed. The idea here is to develop a preprocessing technique to reduce the conflict between the evidences and thus, improve the performance of the DST. This is achieved by assigning a weight to each evidence based on its conflict with other evidences. On the other hand, an evidence with more support from other evidences shows less conflict with them and thus will receive more weight (and vice versa). To evaluate the discrepancy between the evidences, BJS divergence weighted by disagreement degree is used. Reciprocal of this metric weighted by information volume is used as the credibility of the evidences. It is then used to modify the evidences before the combination.

let $\mathfrak{E} = \{E_1, \ldots, E_k, \ldots, E_{n_c}\}$ be the frame of discernment in which the BOEs with BBA $m_i$, $i = 1,2, \ldots, N_c$ are defined. Then to prepare the BOEs for combinations the following steps should be taken.

Step 1. Average belief divergence

In this step, by using BJS (10), for each BOEs the average difference with the other BOEs are assessed as follows,

$$aBJS_i = \frac{1}{N_c - 1} \sum_{\substack{j=1, \\ j \neq i}}^{N_c} BJS(m_i, m_j) \quad (23)$$

Step 2. Disagreement degree

In this step, Eq. (16) is employed to assess the disagreement between the BOEs based on the Jousselme's distance. This is written here again for the sake of convenience,

$$m_{\Delta E, \sim i}^* = 0.5 + \frac{1}{\pi} \arctan\left(\frac{SW - SW_{\sim i}}{\sigma}\right) \quad (24)$$

Step 3. Support degree

In this step, for each BOEs, to assess the support degree the two above measures are combined and inversed. That is,

$$SD_i = \left(aBJS_i \times m_{\Delta E, \sim i}^*\right)^{-1} \quad (25)$$

Step 4. Normalized support degree

Here, the support degrees are normalized such that for each sample they are summed up to unity as follows,

$$\overline{SD} = \frac{SD}{\sum_{i=1}^{N_c} SD_i} \quad (26)$$

Step 5. Credibility degree

To obtain a credibility degree for the BOEs, the $\overline{SD}$ weighted by $\exp(E_d)$ in which $E_d$ is the Deng entropy (9), is used.

$$CD = \exp(E_d) \times \overline{SD} \quad (27)$$

Step 6. Normalized Credibility degree

Here, for each sample, the credibility degree is summed up to unity.

$$\overline{CD} = \frac{CD}{\sum_{i=1}^{N_c} CD_i} \quad (28)$$

Step 7. Weighted evidences

Now, each BOEs are weighted by the normalized credibility as,

$$\boldsymbol{WE}_i = \overline{CD} \times m_i \tag{29}$$

Step 8. Combination

In the last step, the $\boldsymbol{WE}$s are combined based on Dempster's rule (5) to obtain the final output.

$$\widetilde{Y} = \bigoplus_{i=1}^{n_c} WE_i \tag{30}$$

To illustrate the fusion performance, a mathematical example is provided in Appendix I. The proposed fusion method is presented in Algorithm 1.

---
Algorithm 1. The proposed classifier fusion framework
---

**Inputs:**

    $X \in \mathbb{R}^{n_s \times n_f}$: input matrix,

    $T \in \mathbb{R}^{n_s \times n_c}$: target matrix,

    $C_i, i = 1,2, \dots, N_c$: classifiers

**for** $C_i = 1$ **to** $N_c$ **do**        *Classifier generation*

  Train classifier $C_i$

  Evaluate $\widehat{Y}_i = C_i(X) \in \mathbb{R}^{n_s \times n_c}$

  Evaluate $w_i$ using either input of Table 1

  **for** $k = 1$ **to** $n_c$ **do**        *BOE generation*

       Evaluate $m_i(E_k)$ using Eq. (20)

  Evaluate $m_i(\mathfrak{E})$ using Eq. (21)

$\widetilde{M}_i = [m_i(E_k) \; m_i(\mathfrak{E})] \in \mathbb{R}^{n_s \times (n_c+1)}$

**for** $i = 1$ **to** $n_s$ **do**

  Evaluate $aBJS_i$ using Eqs. (23) and (10)

  Evaluate $m^*_{\Delta E, \sim i}$ using Eq. (24)

  Evaluate $\overline{SD}$ using Eqs. (25) and (26)        *Classifier fusion*

  Evaluate $\overline{CD}$ using Eqs. (9), (27), and (28)

  Evaluate $\boldsymbol{WE}_i$ using Eq. (29)

  Combine the results $\widetilde{y}_i$ using Eqs. (5), (8), and (30)

**Output:** $\widetilde{Y} = [\widetilde{y}_1, \dots, \widetilde{y}_i, \dots \widetilde{y}_{n_s}]^T \in \mathbb{R}^{n_s \times (n_c+1)}$

# 4   Application

In this section, the proposed classifier fusion method is first applied to fifteen well-known machine learning datasets. It is finally applied to a vibrational response dataset which was obtained on a set of equiax Polycrystalline Nickel alloy first-stage turbine blades having a range of defect conditions.

## 4.1   Introduction

For each dataset, 12 classifiers from the five methods have been selected to generate the pool of classifiers. They are:

- (i)    $k$NN with $k$ = 5, 7, 9, 11, 13, 15
- (ii)   SVDD with "Gaussian" kernel and width parameters $\sigma$ of 1,2, and 5. they are indicated respectively by SD1, SD2, and SD5. It should be mentioned that although the method has been originally developed as a one-class classifier[5], it can be extended to multi-class classifiers as presented in [29].
- (iii)  SVM with "Gaussian" kernel.
- (iv)   NN with one hidden layer,10 neurons, and "tansig" activation function.

To train the classifiers, each dataset has been randomly divided into three parts: 50% for training, 15% for validation, and 35% for testing. The classifiers have been trained by using the training dataset together with 10-fold cross-validation.

To combine the classifiers, proper classifiers should be selected. However, having in mind that different fusion methods require different numbers and types of the constituent models to present their best performance and also to provide a fair comparison between the fusion techniques, all possible combinations between the trained classifiers have been investigated, i.e. $\binom{12}{i}$ for $i = 1:12$. The number of fusion models that can be obtained by using different numbers of individual models is presented in Table 2. That indicates the total number of fusion models investigated for each dataset is 4095 models.

Table 2. The number of possible fusion models (#Fus) can be obtained by using a different number of individual models (#Idvs)

| #Idvs | 1  | 2  | 3   | 4   | 5   | 6   | 7   | 8   | 9   | 10 | 11 | 12 | Total |
|-------|----|----|-----|-----|-----|-----|-----|-----|-----|----|----|----|-------|
| #Fus  | 12 | 66 | 220 | 495 | 792 | 924 | 792 | 495 | 220 | 66 | 12 | 1  | 4095  |

The proposed fusion method has been applied to the BOEs generated by using six different weightings including the unweighted version, see Table 1. To select one BOEs among the six, the performance of their associated fusion has been assessed on the validation dataset and the one with the highest accuracy has been chosen. It is called "best $P_i$s" in the following. It should be emphasized that, in this study, the main purpose of the validation dataset is to select a proper weight for BOEs and not to avoid overfitting of the models to the training datasets. On the other hand, They have not been used during the training phase.

To highlight the performance of the fusion method in enhancing the classification accuracy, its outcome has been compared with 4 state-of-the-art DST-based fusion methods. These methods have been applied only to the unweighted BOEs. For the sake of abbreviation, these methods are shown by $M_1$[27], $M_2$[33], $M_3$[26], and $M_4$[29].

## 4.2 Benchmark datasets from literature

In this section, the method has been applied to fifteen machine learning datasets collected from UCI [30] and KEEL [31]. Since the main interest of the authors is the development of a framework for non-destructive inspection of complex metal parts (see next section), only datasets with two classes, i.e. $n_c = 2$, but with various imbalance ratios ($IR$) have been selected. The detailed descriptions of these datasets are presented in Table 3. The accuracy of the trained classifiers is assessed on the test dataset (see Table 4). In each dataset, the most accurate models are bold and underlined. Table 5 presents the maximum accuracy among all 4095 models that can be obtained by using different fusion methods over the test dataset. In this table, "NO" stands for "Number of Occurrences" which indicates the number of models among the 4095 models at which the maximum accuracy occurred regardless of the number of individual models used for the fusion. In other words, it indicates the possibility of having the fusion model with the indicated accuracy assuming no prior information is available about selecting the classifiers. Further, the "Sig win" row indicates the number of times that each method gives the maximum classification accuracy.

The table is divided into two sections: unweighted, weighted. In the unweighted section, the BOEs have been generated by using the weight $w_0$ whereas, in the weighted section the weights $w_1, w_2, ..., w_5$ have been utilized to generate the BOEs (see Table 1 for the weights). In each section, the maximum achieved accuracy is shown in bold. For the cases that more than one fusion technique gave the same accuracy, the maximum NO is also bolded. The method with the highest accuracy and the highest NO is underlined.

By comparing different methods in the unweighted section of the table, the proposed method gives the maximum accuracy in 10 out of 15 cases. Moreover, by taking into account the NO to select the best fusion method, the bold and underlined cases, the $P_0$ wins in seven cases, $M_1$ in one case, and the other methods never win. By performing the same comparison among different versions of the proposed methods, i.e. $P_0$ to $P_5$, one can see that $P_5$ gives the maximum accuracy in 12 cases out of 15 whereas, the other weightings show poorer performances than that of the unweighted version. An interesting observation is that for the cases that $P_0$ did not give the highest accuracy, at least one of its variations gave that accuracy. This could justify the simultaneous use of all six versions of the proposed method and use a validation dataset to select the best one instead of the preselection of one weighting to generate BOEs. This thus leads to the exceptional performance of the "best $P_i$s" which gives the maximum test accuracy in 14 out of 15 cases. The only case that it did not give the highest accuracy, case 13, it can be seen that it gave the accuracy equal to the other fusion methods.

As the last analysis, the computational time of different fusion approaches has been compared. Table 6 shows the result for the case of having 11 constituent models in the ensemble. The results were obtained by the command "*cputime*" in Matlab. The results indicate that the $M_2$ is the most efficient method but in the sense of accuracy, it showed the poorest performance among the considered methods. All in all, it can be concluded using the proposed approach leads to more accurate results with the expense of negligible extra more computation time.

Table 3. summary of the UCI and KEEL datasets. Here $n_s$ is the number of samples, $n_f$ is the number of features and $IR$ is the imbalance ratio.

| Label | Name | $n_s$ | $n_f$ | IR |
|---|---|---|---|---|
| 1 | Crab | 200 | 6 | 1.00 |
| 2 | Ovarian Cancer | 216 | 63 | 1.27 |
| 3 | Ionosphere | 351 | 33 | 1.78 |
| 4 | Wisconsin | 683 | 9 | 1.86 |
| 5 | Pima | 768 | 8 | 1.87 |
| 6 | Breast Cancer | 699 | 9 | 1.9 |
| 7 | Haberman | 306 | 3 | 2.78 |
| 8 | Vehicle 2 | 846 | 16 | 2.88 |
| 9 | Glass | 214 | 9 | 3.20 |
| 10 | Yeast 3 | 1484 | 8 | 8.10 |
| 11 | Ecoli 4 | 336 | 7 | 15.8 |
| 12 | Abalone 9-18 | 731 | 8 | 16.4 |
| 13 | Yeast 4 | 1484 | 8 | 28.1 |
| 14 | Yeast 6 | 1484 | 8 | 41.4 |
| 15 | Abalone 19 | 4174 | 7 | 129.44 |

Table 4. Accuracy (in percent) of different classification methods applied to UCI and KEEL datasets, assessed on the test set.

|    | LDA   | 5NN   | 7NN   | 9NN   | 11NN  | 13NN  | 15NN  | SD1   | SD2   | SD5   | SVM   | NN    |
|----|-------|-------|-------|-------|-------|-------|-------|-------|-------|-------|-------|-------|
| 1  | 94.00 | **96.00** | **96.00** | **96.00** | **96.00** | **96.00** | **96.00** | 95.00 | 88.00 | 50.00 | 94.00 | **96.00** |
| 2  | 81.65 | 91.74 | 91.74 | 91.74 | 91.74 | 91.74 | 91.74 | 55.96 | 67.88 | 88.99 | 93.57 | **94.49** |
| 3  | 82.95 | 83.52 | 82.38 | 81.81 | 81.25 | 80.11 | 80.68 | 65.90 | 88.06 | 88.63 | 92.04 | **92.61** |
| 4  | 92.88 | 88.70 | 88.28 | 88.70 | 88.70 | 87.86 | 87.86 | 92.88 | 93.30 | 92.05 | 94.56 | **98.74** |
| 5  | 78.13 | 67.71 | 66.67 | 67.71 | 66.67 | 65.10 | 64.58 | 65.10 | 65.10 | 65.10 | 78.64 | **86.46** |
| 6  | 94.57 | 80.85 | 88.28 | 88.57 | 89.42 | 88.28 | 88.57 | 93.14 | 95.14 | 95.14 | 95.71 | **97.71** |
| 7  | 75.70 | 73.83 | 71.02 | 72.90 | 73.83 | 73.83 | 73.83 | 73.83 | 68.22 | 67.29 | 74.77 | **79.44** |
| 8  | 96.62 | 94.93 | 94.25 | 93.91 | 93.58 | 93.58 | 92.56 | 79.39 | 93.24 | 85.47 | 96.28 | **99.32** |
| 9  | 92.59 | 93.51 | 93.51 | 93.51 | 92.59 | 91.66 | 91.66 | 88.88 | 95.37 | 75.92 | 93.51 | **98.15** |
| 10 | 94.21 | 94.02 | 93.64 | 93.64 | 93.83 | 94.21 | 94.02 | 91.32 | 90.55 | 90.17 | 92.29 | **95.18** |
| 11 | 98.30 | **99.15** | **99.15** | **99.15** | **99.15** | **99.15** | 96.61 | 94.92 | 98.30 | 5.93  | **99.15** | **99.15** |
| 12 | **96.09** | 94.14 | 94.14 | 93.35 | 93.75 | 94.53 | 94.14 | 93.75 | 90.62 | 5.86  | 93.35 | **96.09** |
| 13 | 97.12 | 96.62 | 96.73 | 96.92 | 96.53 | 96.53 | 96.73 | 96.53 | 84.42 | 80.38 | 96.34 | **97.50** |
| 14 | 96.72 | 97.30 | **97.88** | **97.88** | 97.49 | 97.49 | 97.68 | **97.88** | 97.49 | 2.31  | 97.68 | **97.88** |
| 15 | 98.69 | 99.24 | 99.24 | 99.24 | 99.24 | 99.24 | 99.24 | 97.05 | 93.85 | 0.75  | 99.24 | **99.31** |

Table 5. Maximum Accuracy (in percent) on the test set of UCI and KEEL datasets obtained through all possible combinations of classifiers.

|   |   | Unweighted | | | | | Weighted | | | | | |
|---|---|---|---|---|---|---|---|---|---|---|---|---|
|   |   | $M_1$ | $M_2$ | $M_3$ | $M_4$ | $P_0$ | $P_1$ | $P_2$ | $P_3$ | $P_4$ | $P_5$ | best $P_i$s |
| 1 | Acc | **98.00** | **98.00** | **98.00** | 97.00 | **98.00** | 98.00 | 98.00 | 98.00 | 98.00 | 98.00 | <u>98.00</u> |
|   | NO | 9 | 3 | **42** | 151 | 31 | 69 | **148** | 44 | 45 | 77 | <u>154</u> |
| 2 | Acc | 94.50 | 95.41 | 96.33 | **97.25** | 97.25 | 97.25 | 97.25 | 97.25 | 97.25 | 97.25 | <u>97.25</u> |
|   | NO | 8 | 47 | 47 | 5 | **59** | 9 | 6 | **22** | 9 | 6 | <u>71</u> |
| 3 | Acc | 93.75 | **96.02** | **96.02** | **96.02** | 95.45 | 96.02 | 96.02 | 96.02 | 96.02 | 96.02 | <u>96.02</u> |
|   | NO | 4 | 4 | **27** | 15 | 4 | **51** | 8 | **51** | 6 | 2 | <u>53</u> |
| 4 | Acc | 98.75 | 98.75 | 98.75 | 98.75 | <u>**99.16**</u> | 98.75 | 98.75 | 98.75 | 98.75 | 98.75 | <u>99.16</u> |
|   | NO | 1 | 1 | 6 | 16 | <u>**7**</u> | 8 | 7 | 11 | 7 | 7 | <u>7</u> |
| 5 | Acc | 86.97 | 86.45 | 86.98 | **87.50** | **87.50** | 86.97 | **87.50** | 86.97 | **87.50** | **87.50** | <u>87.50</u> |
|   | NO | 4 | 1 | 2 | 3 | **69** | 11 | 1 | 7 | 1 | 1 | <u>70</u> |
| 6 | Acc | 97.71 | 98.00 | 98.57 | 98.57 | <u>**98.86**</u> | 98.57 | 98.57 | **98.86** | 98.57 | 98.57 | 98.86 |
|   | NO | 5 | 3 | 3 | 12 | <u>**6**</u> | 20 | 73 | **1** | 94 | 83 | 5 |
| 7 | Acc | <u>**81.31**</u> | 79.44 | **81.31** | 80.37 | 80.37 | **81.31** | 80.37 | **81.31** | 80.37 | **81.31** | <u>81.31</u> |
|   | NO | <u>**2**</u> | 3 | 1 | 1 | 4 | 1 | 4 | **2** | 4 | 1 | <u>2</u> |
| 8 | Acc | **99.66** | 99.32 | **99.66** | **99.66** | <u>**99.66**</u> | 99.66 | 99.66 | 99.66 | 99.66 | 99.66 | 99.66 |
|   | NO | 1 | 1 | 1 | 21 | <u>**73**</u> | 13 | **31** | 29 | 18 | 20 | 68 |
| 9 | Acc | **98.15** | **98.15** | **98.15** | **98.15** | **98.15** | 98.15 | 98.15 | 98.15 | 98.15 | 98.15 | 98.15 |
|   | NO | 2 | 2 | 10 | 34 | <u>**276**</u> | 25 | 4 | 24 | 6 | 8 | 258 |
| 10 | Acc | 95.19 | 95.38 | 95.38 | 95.18 | <u>**95.95**</u> | 95.76 | 95.38 | 95.56 | 95.38 | **95.95** | <u>95.95</u> |
|   | NO | 4 | 1 | 1 | 14 | <u>**3**</u> | 2 | 6 | 5 | 1 | **2** | <u>3</u> |
| 11 | Acc | 99.15 | 99.15 | **100** | 99.15 | 99.15 | <u>**100**</u> | 99.15 | **100** | 99.15 | 99.15 | 100 |
|   | NO | 48 | 1592 | **2** | 1884 | 2114 | <u>**6**</u> | 2069 | **3** | 2005 | 2073 | 5 |
| 12 | Acc | **96.88** | 96.09 | 96.48 | 96.09 | 96.48 | 96.48 | **96.88** | **96.88** | **96.88** | **96.88** | 96.88 |
|   | NO | 1 | 4 | 2 | 11 | 230 | 4 | **1** | **1** | **1** | **1** | 1 |
| 13 | Acc | 97.89 | 97.89 | 97.89 | 97.89 | <u>**98.01**</u> | 97.89 | 97.89 | 97.69 | 97.69 | **98.01** | 97.89 |
|   | NO | 1 | 4 | 2 | 1 | <u>**4**</u> | 13 | 8 | 18 | 12 | **2** | 8 |
| 14 | Acc | 98.07 | 98.27 | **98.65** | 98.46 | <u>**98.65**</u> | **98.65** | 98.27 | 98.27 | 98.27 | **98.65** | <u>98.65</u> |
|   | NO | 20 | 2 | 3 | 1 | <u>**29**</u> | 3 | 7 | 3 | 1 | **7** | <u>29</u> |
| 15 | Acc | 99.31 | 99.31 | **99.39** | **99.39** | 99.31 | **99.39** | 99.31 | 99.31 | **99.39** | <u>**99.39**</u> | <u>99.39</u> |
|   | NO | 4 | 1 | **2** | 1 | 2 | 2 | 2 | 2 | 1 | <u>**4**</u> | <u>4</u> |
| Sig win | | 5 | 3 | 8 | 6 | 10 | 9 | 6 | 9 | 8 | 12 | 14 |

Table 6. Computational time (in second) of different fusion methods applied to the UCI and KEEL benchmarks. It is reported for the ensemble of 11 constituent models.

| Dataset | $M_1$ | $M_2$ | $M_3$ | $M_4$ | $P_0$ |
|---|---|---|---|---|---|
| 1 | 0.0379 | 0.0112 | 0.0179 | 0.0491 | 0.0379 |
| 2 | 0.0335 | 0.0112 | 0.0134 | 0.0313 | 0.0335 |
| 3 | 0.0535 | 0.0245 | 0.0246 | 0.0558 | 0.0625 |
| 4 | 0.1138 | 0.0312 | 0.0379 | 0.1071 | 0.1004 |
| 5 | 0.1383 | 0.0558 | 0.0491 | 0.1138 | 0.1049 |
| 6 | 0.1116 | 0.0401 | 0.0379 | 0.1161 | 0.0982 |
| 7 | 0.0535 | 0.0156 | 0.0245 | 0.0535 | 0.0580 |
| 8 | 0.1294 | 0.0491 | 0.0535 | 0.1406 | 0.1272 |
| 9 | 0.0334 | 0.0134 | 0.0290 | 0.0334 | 0.0312 |
| 10 | 0.2544 | 0.0915 | 0.0915 | 0.2500 | 0.2812 |
| 11 | 0.0513 | 0.0223 | 0.0245 | 0.0536 | 0.0603 |
| 12 | 0.1205 | 0.0558 | 0.0603 | 0.1428 | 0.1049 |
| 13 | 0.2276 | 0.0781 | 0.0826 | 0.2232 | 0.2349 |
| 14 | 0.2566 | 0.0714 | 0.0871 | 0.2612 | 0.2143 |
| 15 | 0.6328 | 0.2006 | 0.3138 | 0.7482 | 0.5251 |
| mean | 0.1499 | 0.0514 | 0.0631 | 0.1586 | 0.1383 |

## 4.3 Vibrational dataset of first-stage turbine blades

In this section, the proposed classifier fusion method has been applied to an experimental dataset. The broadband vibrational response has been acquired for equiax Polycrystalline Nickel alloy first-stage turbine blades with complex geometry and various damage features. Two views of CAD model of the blade are shown in Figure 3. The cooling channel in the middle can be seen in the transparent view. The damages in the blades range from microstructure changes due to over-temperature, airfoil cracking, inter-granular attack (corrosion), to thin walls due to casting, MRO operations, and/or service wear. It should be emphasized that prior to apply any classification methodology, the health condition of the samples has been determined by some means, e.g. X-ray, visual testing, penetrant testing, ultrasonics, operator experience, etc., and that knowledge is used to train and evaluate the performance of the proposed classification approach.

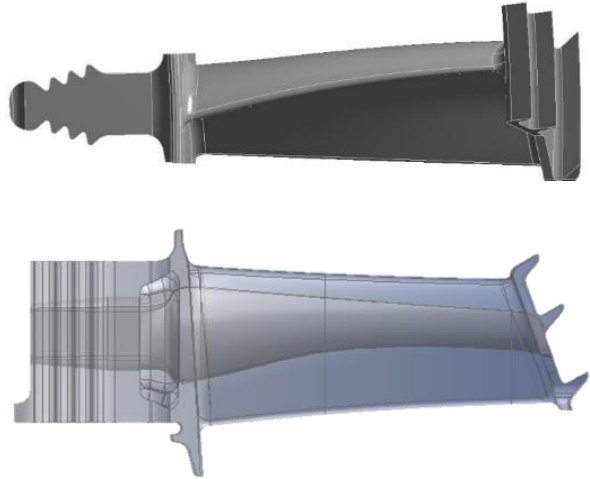

Figure 3. Two views of the CAD model of the Equiax Polycrystalline Nickel alloy first-stage turbine blade. Bottom plot shows a transparent view to illustrate the cooling channel.

By using one actuator and two sensors, i.e. Single Input Multiple Output SIMO, the amplitude of the frequency response function (FRF) has been collected from each blade in the range of [3, 38] kHz. From the FRFs, 15 frequencies, $F_i\ with\ i = 1,2,\dots,15$, are extracted to be used as the features for the classifiers. By measuring the FRF from 192 healthy and 33 defected blades, the dataset has been created with $IR = 5.82$. To assess the performance of different classifiers and fusion techniques, besides accuracy Eq. (17), their precision, Eq. (18) in detecting defected samples has also been reported. In two-class problems, it is referred to as the Specificity (Spc). A schematic of the experimental test procedure is shown in Figure 4.

In the following, three different analyses have been carried out:

(i) Single model analysis: elaboration of one random train/validation/test dataset and the associated individual classifiers and their effects on the fusion models

(ii) Statistical analysis: investigation on the effect of different train/validation/test datasets on the performance of individual classifiers and their fusion models

(iii) Noise analysis: investigation on the effect of noise in the measurement data on the performance of the fusion models

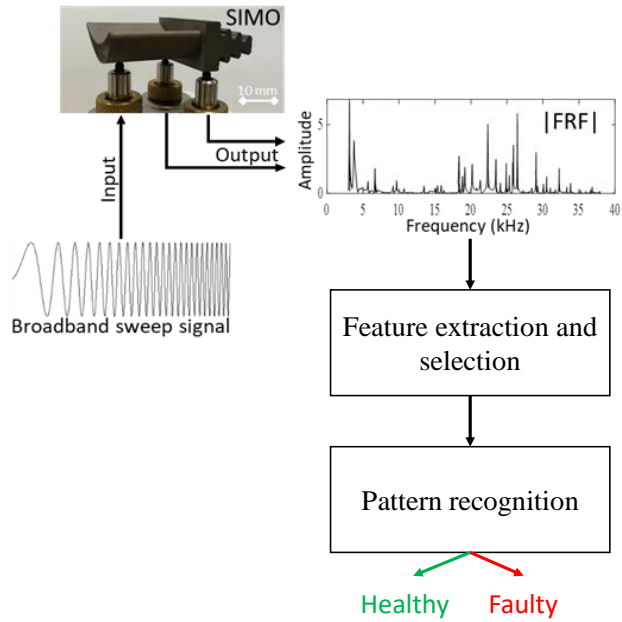

Figure 4. Schematic of the test procedure with one actuator and two sensors. Figure reproduced from [8].

### 4.3.1 Single model analysis

The accuracy and the specificity of the trained classifiers have been assessed on the test dataset, and are listed in Table 7. These classifiers were combined to generate 4095 fusion models, see Table 2. The performance of each of the fusion models in the sense of accuracy and specificity have been assessed on the test dataset. The average and standard deviation of the performances have been obtained over the sets of fusion models with the same number of constituent models. Figure 5 and Figure 6 show this result for different fusion methods and for the sets of fusion models with 2, 4, 8, and 12 constituent models. The full results are provided as supplementary information.

Figure 5a and Figure 5b show the average accuracy and specificity of the proposed fusion technique together with different weightings for generating BOEs, see Table 1. The error bars denote the standard deviation. Here, the "best $P_i$s" is shown by the red bar. It can be observed that in this application, generating BOEs with weighing $w_3$, in most cases, leads to the best average performance on the test dataset. This will be further investigated in the following.

Figure 6 shows the same results for the fusion methods from the literature. In order to compare their performance with the proposed method, the "best $P_i$s" in Figure 5 is shown here by "Prop". As can be seen, the proposed method shows higher average accuracy and specificity than the other methods. Moreover, improving the average specificity by including more models can be observed.

Table 7. Performance of the trained classifiers assessed on the test dataset

| No. | Method | Train Acc. (%) | Valid Acc (%) | Test Acc (%) | Test Spc (%) |
|---|---|---|---|---|---|
| 1 | LDA | 97.32 | 94.12 | 96.25 | 83.33 |
| 2 | 5NN | 90.18 | 91.18 | 90.00 | 41.66 |
| 3 | 7NN | 90.18 | 94.12 | 87.50 | 25.00 |
| 4 | 9NN | 91.07 | 94.12 | 87.50 | 25.00 |
| 5 | 11NN | 90.18 | 91.18 | 87.50 | 16.66 |
| 6 | 13NN | 89.29 | 85.30 | 86.25 | 8.33 |
| 7 | 15NN | 89.29 | 85.30 | 85.00 | 0.00 |
| 8 | SD1 | 100 | 85.30 | 85.00 | 0.00 |
| 9 | SD2 | 100 | 85.30 | 85.00 | 0.00 |
| 10 | SD5 | 98.21 | 94.12 | 95.00 | 100 |
| 11 | SVM | 98.21 | 94.12 | 98.75 | 100 |
| 12 | NN | 97.32 | 94.12 | 95.00 | 75.00 |

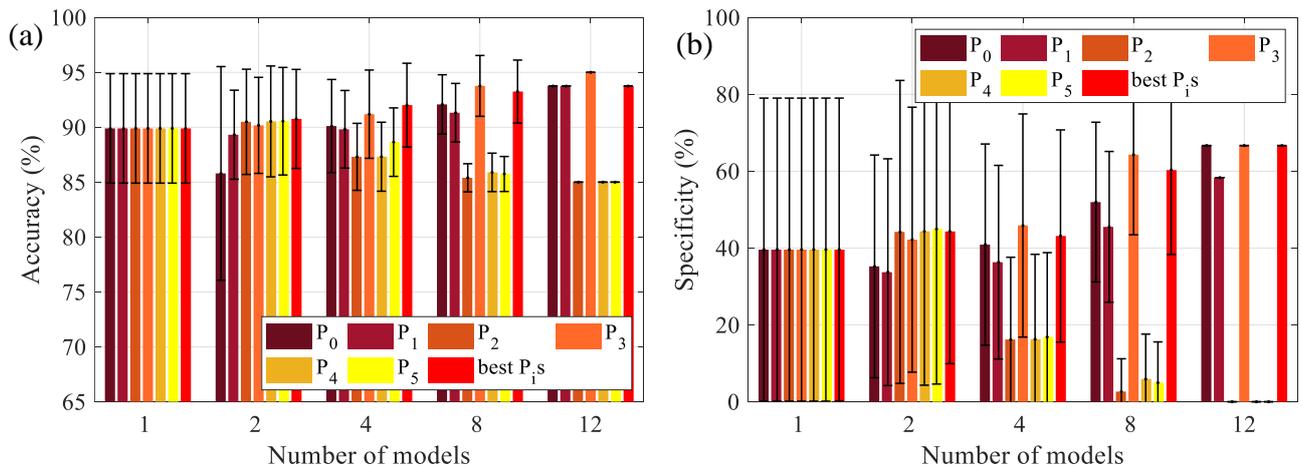

Figure 5. Average accuracy (a) and specificity (b) of the proposed fusion method over combinations made by the same number of constituent models. Different weightings used for the BOEs are shown in different colors. best $P_i$s stands for the model with the best validation accuracy. The error bars indicate standard deviation

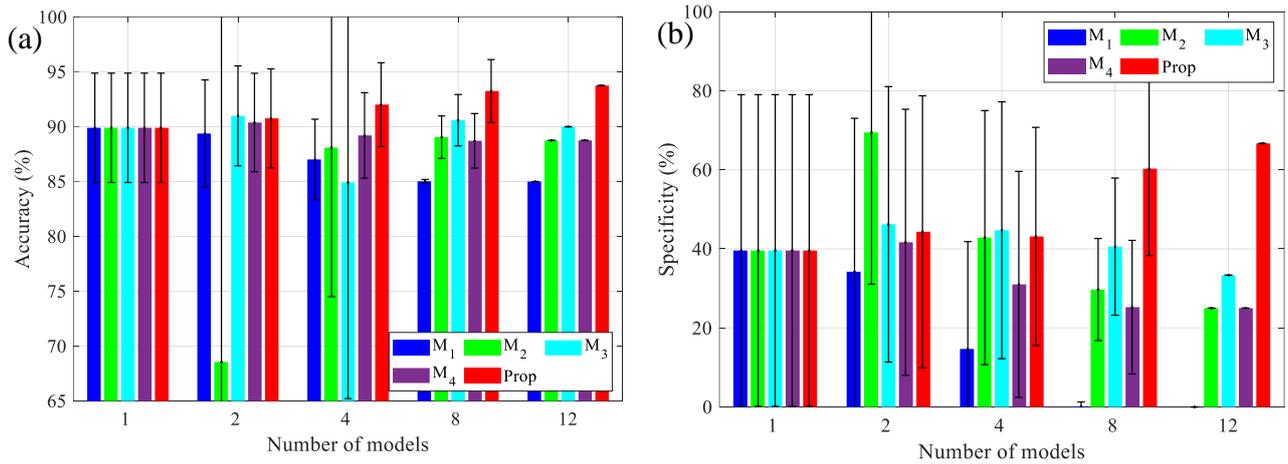

Figure 6. Average accuracy (a) and specificity (b) of the fusion method over combinations made by the same number of constituent models. The "Prop" is the same as the "best $P_i$s" in Figure 5. The error bars indicate standard deviation.

For further comparison between the methods, an investigation has been made on the *best* model obtained by each approach. The *best* model is defined as the most accurate model among the fusion models obtained by the same method and with the same number of constituent models. Figure 7 shows the performance of these *best* models versus the number of constituent models. The effect of different weightings can also be inferred. It can be observed that the best performance of the *best* models occurs in $P_4$ and $P_5$ for the ensemble of two and four constituent models respectively. That means, although $P_4$ and $P_5$ show promising result in this case, no general conclusion can be drawn for the pre-selection of weightings if the best performance is of interest. This could justify the use of a validation dataset to select the "best $P_i$s" among different $P_i$s. Figure 7b shows the specificity associated with the *best* models. In cases with several *best* models, the highest specificity is depicted.

Figure 8 shows the *best* performance of the other fusion methods to compare with that of the proposed method. Here, the result of the proposed method is shown by "Prop" and it is the same as the "best $P_i$s" in Figure 7. It can be seen that the *best* fusion model made by the proposed approach could outperform the *best* constituent model and reaches 100% accuracy by combining two and four models. For $M_2$ this could also occur only when two models have been combined. Moreover, it can be inferred that including more models into the ensemble could deteriorate the performance of all fusion methods. This illustrates the importance of proper classifier selection for fusion.

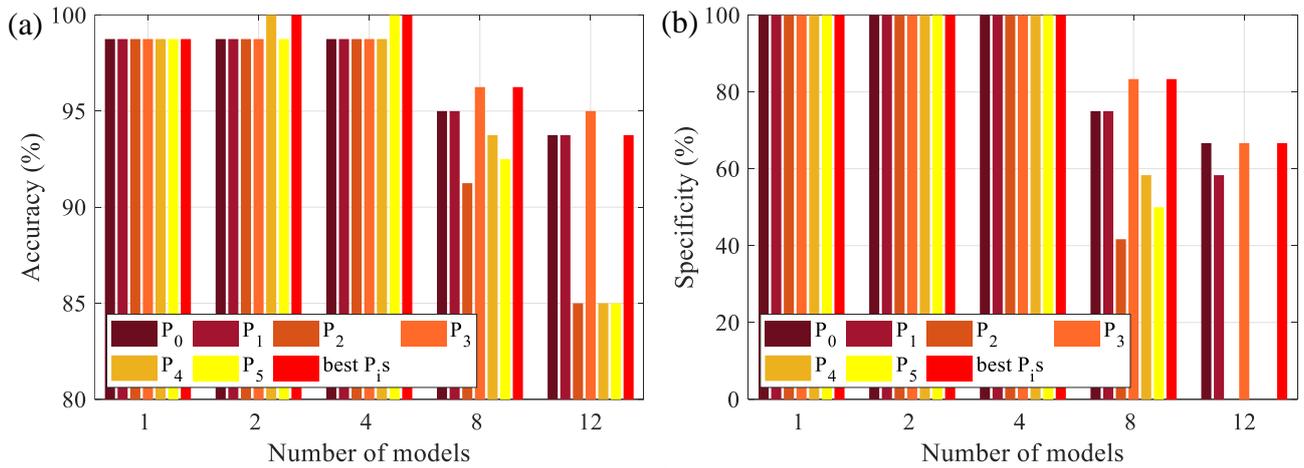

Figure 7. Accuracy (a) and specificity (b) of the best model generated by the proposed fusion method with the same number of constituent models. Different weightings used for the BOEs are shown in different colors. best $P_i$s stands for the model with the best validation accuracy.

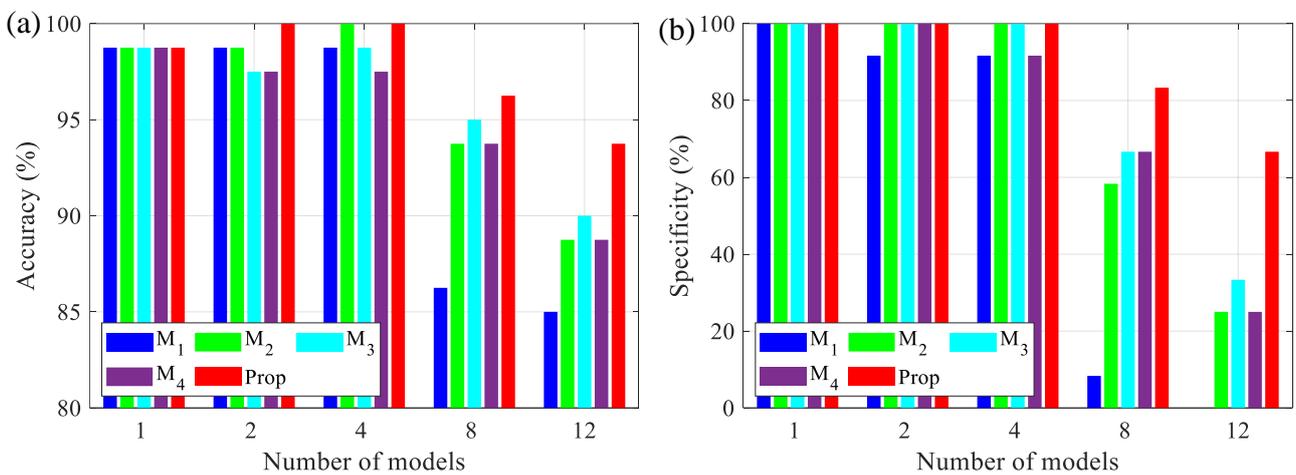

Figure 8. Average accuracy (a) and specificity (b) of the best model generated fusion method over combinations made by the same number of constituent models. The "Prop" is the same as the "best $P_i$s" in Figure 5.

*4.3.2 Statistical analysis*

In order to provide a thorough comparison, a statistical analysis has been devised. For this purpose, the whole procedure has been repeated 50 times. The average performance of the individual classifiers is shown in Figure 9. In this figure, the error bars indicate the standard deviation associated with each method. It can be seen that the SVM method shows the best performance among all the methods.

To provide statistics for the fusion methods, at each iteration the *best* model, as defined in the previous section, is selected. Here, to reduce the computational burden, the analysis has been made up to five constituent models. The results are shown in Figure 10. The bars indicate the average performance of each method and the error bars show their associated standard deviation. At each iteration, the proposed method has been used in conjunction with all six weighings, and the model with the best performance on the validation dataset is selected, which is shown here as the "Prop".

It can be seen that the proposed method outperforms the other methods in all cases. Besides, the best performance of the "Prop" is obtained when there are two or three constituent models available for combination.

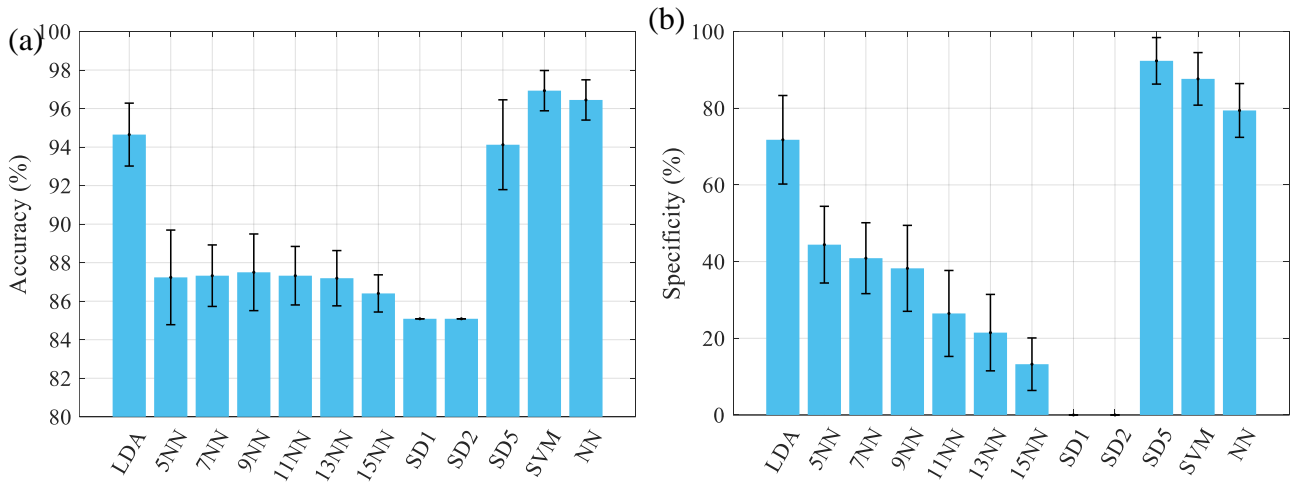

Figure 9. Average accuracy (a) and specificity (b) of the individual models over 50 repetitions of data resampling and classifier training. Performances have been assessed on the test dataset. The error bars indicate the associated standard deviation.

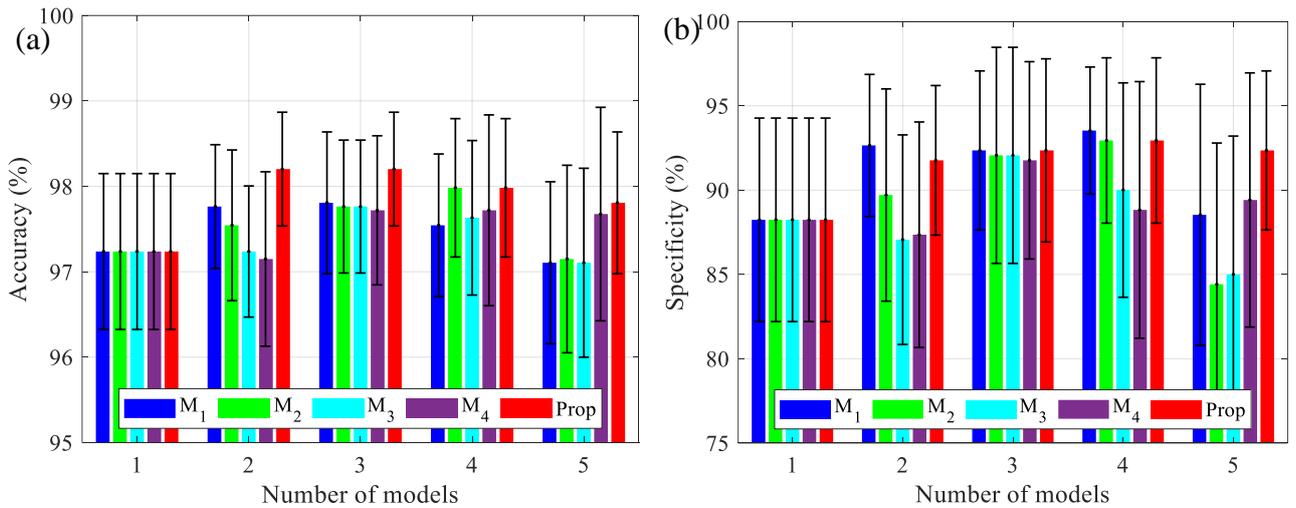

Figure 10. Average accuracy (a) and specificity (b) of the *best* models over 50 repetitions of data resampling and classifier training. Performances have been assessed on the test dataset. The error bars indicate the standard deviation.

### 4.3.3  Noise analysis

Experimental measurement includes different levels of noise and/or error stemming from the environment, the instruments, or the operator. Hence, a proper classification method should be able to remain accurate and stable in the presence of such noise. To this end, the features estimated from the FRFs have been polluted with different levels of noise and their effect on the fusion performance has been investigated. In the following the result for 1% RMS noise levels has been presented, the result for 2% and 5% RMS noise is available in Appendix II.

To assess the performance of the fusion techniques, up to five models are combined together. The average of the *best* models obtained by each approach over the 50 repetitions is presented in Figure 11. The results indicate that the proposed method outperforms the other methods and also the constituent models. In this figure, it can be observed that the best fusion performance can be obtained by combining two models. The result for higher levels of noise in Appendix II shows that combining three and/or four models is more suitable.

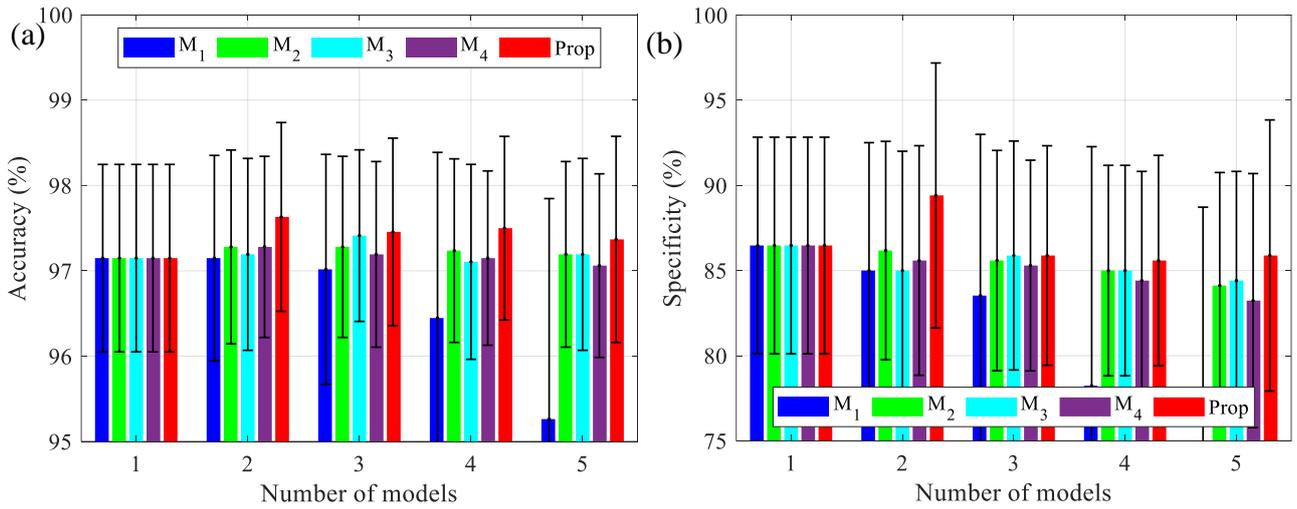

Figure 11. Average accuracy (a) and specificity (b) of the *best* models over 50 repetitions of data resampling and classifier training. Performances have been assessed on the test dataset. The error bars indicate the standard deviation.

### 4.3.4 Overall analysis

In this section, the *best* models obtained by each fusion technique in the presence of different levels of noise have been compared. The results are shown in Figure 12. In this figure, "Idv" stands for individual models which are shown in light blue bars. This bar shows the range of variation in the accuracy of all individual models generated in statistical analysis with one specific noise level.

To generate the other bars, at each repetition of the statistical analysis, the most accurate model among all the fusion models generated by the same fusion technique with one specific noise level, regardless of the number of constituent models, has been selected. The bars show the range of variation in these models. The cross-circle ($\otimes$) illustrates the mean-value of each group.

The figure indicates the superiority of the proposed fusion method in all cases. Especially for higher noise levels. In the case with 5% RMS noise, the most accurate model in the individual model has an accuracy of 93.86% whereas, the most accurate model in the proposed fusion techniques reaches up to 97.37%.

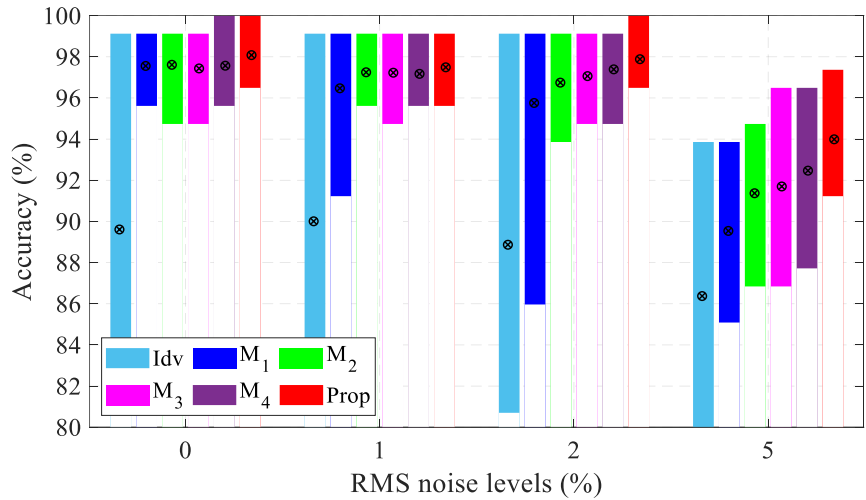

Figure 12. Range of variation in the accuracy of the individual models (Idv) and the best model of the fusion techniques obtained through 50 repetitions versus RMS noise levels.

# 5 Conclusion

In this paper, a multi-classifier information fusion approach has been developed to improve the performance in 2-class classification problems. A new method based on the Dempster-Shafer theory of evidence was introduced for classifier fusion to alleviate the problem of conflicting evidences. The proposed method has been equipped with six different weightings to generate BOEs.

To evaluate the performance of the proposed method, it has been applied to 15 benchmarks from machine learning repositories. A comparison with four DST-based fusion method reveals that the proposed method could outperform the other methods and also the individual classifiers. It was also shown that the best performance of the fusion methods occurs rarely and highly depends on the selected constituent classifiers. This implies the necessity of proper classifier selection for information fusion.

The fusion procedure has also been tested on vibrational data collected from equiax polycrystalline Nickel alloy first-stage turbine blades with complex geometry and different types and severities of damage. Classification accuracy of 100% has been obtained on the test dataset, indicating the high performance of the developed framework. Through statistical analysis with different levels of noise, the robustness of the proposed method was concluded.

# 6 Acknowledgment

The authors gratefully acknowledge the ICON project DETECT-ION (HBC.2017.0603) which fits in the SIM research program MacroModelMat (M3) coordinated by Siemens (Siemens Digital Industries Software, Belgium) and funded by SIM (Strategic Initiative Materials in Flanders) and VLAIO

(Flemish government agency Flanders Innovation & Entrepreneurship). Vibrant Corporation is also gratefully acknowledged for providing anonymous datasets of the blades.

# Appendix I. Example on the fusion method

Suppose there are four individual classifiers ($C_1$, $C_2$, $C_3$, and $C_4$) trained on a three-class classification problem. Given an instance $X$, these classifiers give the evidence vector $\hat{m}_1 = (0.5, 0.1, 0.4)$, $\hat{m}_2 = (0.3, 0.3, 0.4)$, $\hat{m}_3 = (0.5, 0.0, 0.5)$, and $\hat{m}_4 = (0.4, 0.2, 0.4)$. Here $n_c = 2$, $N_c = 4$, and $\mathfrak{E} = \{E_1, E_2, \{E_1, E_2\}\}$. To combine the classifiers based on the proposed fusion technique, the following steps should be taken:

Step 1. To obtain the average BJS of the BOEs, take $m_1$ and $m_i, i = 1, \ldots, 4$, $BJS(m_1, m_i) = [0.0, 0.056, 0.054, 0.0163]$ thus $aBJS_1 = 0.042$. Do the procedure for $m_i, i = 1, \ldots, 4$ leads to $aBJS = [0.042, 0.080, 0.111, 0.046]$.

Step 2. The Disagreement associated with each BOEs are obtained as follows,

I.  $\bar{m} = [0.425, 0.150, 0.425]$, $d_J(m_i, \bar{m}) = [0.094, 0.197, 0.184, 0.061]$ and thus $SW = 0.134$.

II. Take $m_1$, $\bar{m}_{\sim 1} = [0.400, 0.167, 0.433]$ and $d_J(m_i, \bar{m}_{\sim 1}) = [0.170, 0.205, 0.047]$, and thus $SW_{\sim 1} = 0.0141$. Do the procedure for $m_i, i = 1, \ldots, 4$ leads to $SW_{\sim i} = [0.141, 0.099, 0.094, 0.154]$.

III. $m^*_{\Delta E, \sim i} = [0.496, 0.522, 0.525, 0.487]$

Step 3. The support degree of the BOEs are

$$SD = [47.95, 23.87, 17.09, 45.02]$$

Step 4. Normalized support degree is

$$\overline{SD} = [0.358, 0.178, 0.128, 0.336]$$

Step 5. The credibility of the BOEs are,

I.  $E_d = [0.601, 0.664, 0.540, 0.650]$
II. $CD = [0.653, 0.346, 0.219, 0.643]$

Step 6. Normalized credibility is,

$$\overline{CD} = [0.351, 0.186, 0.118, 0.346]$$

Step 7. The weighted evidences are,

$$WE = \begin{bmatrix} 0.175 & 0.035 & 0.140 \\ 0.056 & 0.056 & 0.074 \\ 0.059 & 0.000 & 0.059 \\ 0.136 & 0.069 & 0.136 \end{bmatrix}$$

Step 8. The final output obtained by Demspter's rule of combination is,

$$\widetilde{Y} = [0.818, 0.1265, 0.056]$$

MaxIndex($[0.818, 0.1265, 0.056]$) = 1, Hence, the instance $X$ belongs to class 1.

# Appendix II.   Noise analysis

In this section, the same results as Section 4.3.3 have been presented for the case that the features were polluted with 2% and 5% RMS noise levels. They are shown respectively in Figure II-1 and Figure II-2. They indicate that by including more models up to four models, the fusion performance is improving.

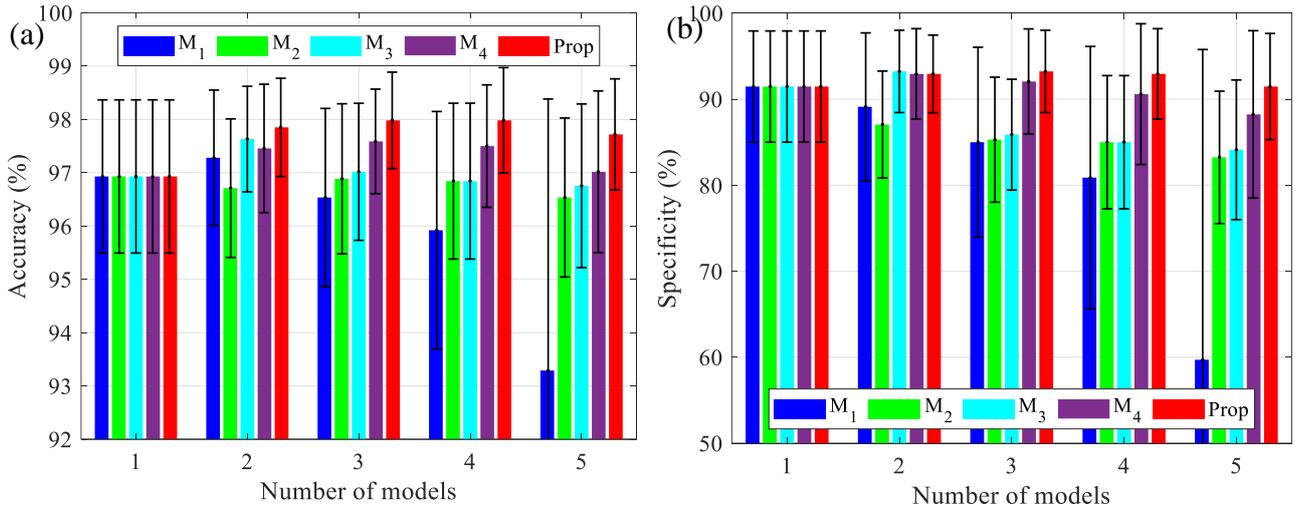

Figure II-1. Average accuracy (a) and specificity (b) of the *best* models over 50 repetitions of data resampling and classifier training. Performances have been assessed on the test dataset. The error bars indicate the standard deviation.

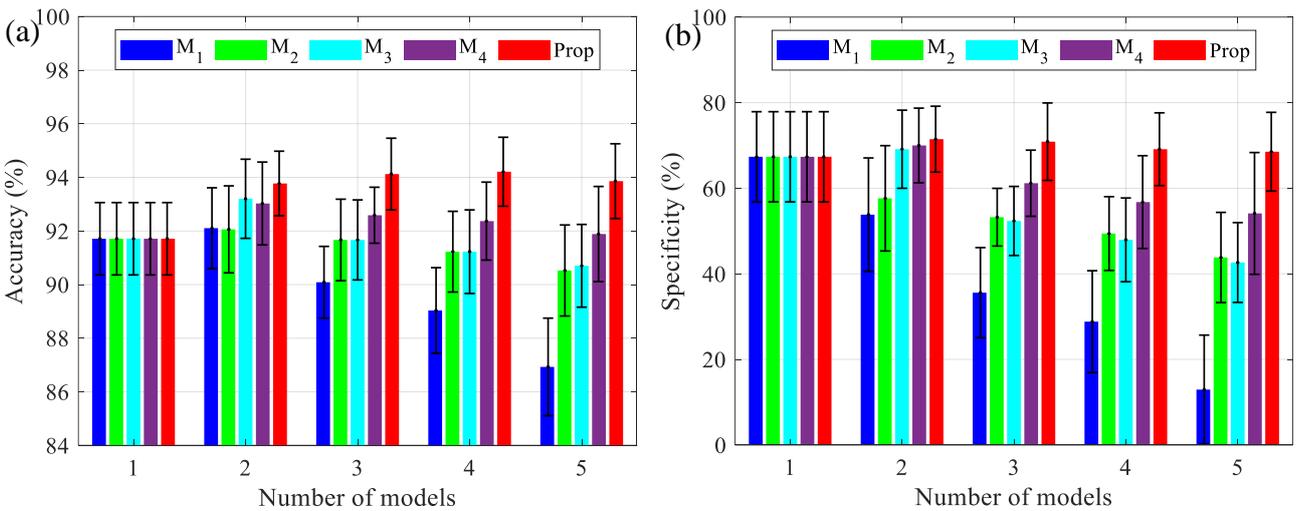

Figure II-2. Average accuracy (a) and specificity (b) of the *best* models over 50 repetitions of data resampling and classifier training. Performances have been assessed on the test dataset. The error bars indicate the standard deviation.